\documentclass{article}

\PassOptionsToPackage{numbers}{natbib}
\usepackage[dblblindworkshop, final]{neurips_2025}

\usepackage[utf8]{inputenc} 
\usepackage[T1]{fontenc}    
\usepackage{hyperref}       
\usepackage{url}            
\usepackage{booktabs}       
\usepackage{amsfonts}       
\usepackage{nicefrac}       
\usepackage{microtype}      
\usepackage{xcolor}         
\usepackage{amsmath}
\usepackage{graphicx}
\usepackage[numbers]{natbib}
\usepackage{todonotes}
\usepackage{array}

\title{Multimodal Bayesian Network for Robust Assessment
of Casualties in Autonomous Triage}
\workshoptitle{Structured Probabilistic Inference \& Generative Modeling}

\author{%
  Szymon Rusiecki$^{1,2}$, Cecilia G. Morales$^2$, Kimberly Elenberg$^2$, Leonard Weiss$^3$,\\ 
  \textbf{Artur Dubrawski$^2$} \\[0.75ex]
  $^1$AGH University of Krakow, Kraków, Poland \\
  $^2$Carnegie Mellon University, Pittsburgh, PA, USA \\
  $^3$University of Pittsburgh, Pittsburgh, PA, USA \\[0.75ex]
  \texttt{rusiecki@student.agh.edu.pl, \{cgmorale, kelenber, awd\}@andrew.cmu.edu,} \\
  \texttt{weissls2@upmc.edu}
}

\begin{document}

\maketitle

\begin{abstract}
Mass Casualty Incidents can overwhelm emergency medical systems and resulting delays or errors in the assessment of casualties can lead to preventable deaths.
We present a decision support framework that fuses outputs from multiple computer vision models, estimating signs of severe hemorrhage, respiratory distress, physical alertness, or visible trauma, into a Bayesian network constructed entirely from expert-defined rules.
Unlike traditional data-driven models, our approach does not require training data, supports inference with incomplete information, and is robust to noisy or uncertain observations.
We report performance for two missions involving 11 and 9 casualties, respectively, where our Bayesian network model substantially outperformed vision-only baselines during evaluation of our system in the DARPA Triage Challenge (DTC) field scenarios.
The accuracy of physiological assessment improved from 15\% to 42\% in the first scenario and from 19\% to 46\% in the second, representing nearly threefold increase in performance.
More importantly, overall triage accuracy increased from 14\% to 53\% in all patients, while the diagnostic coverage of the system expanded from 31\% to 95\% of the cases requiring assessment.
These results demonstrate that expert-knowledge-guided probabilistic reasoning can significantly enhance automated triage systems, offering a promising approach to supporting emergency responders in MCIs.
This approach enabled Team Chiron to achieve 4th place out of 11 teams during the 1st physical round of the DTC.
\end{abstract}    
\section{Introduction}
\label{sec:intro}

When a terrorist attack struck the Boston Marathon in 2013, emergency responders faced around 250 casualties within minutes, overwhelming the city's trauma capacity and forcing life-or-death triage decisions under extreme pressure~\cite{gunaratna2013homegrown}. Such Mass Casualty Incidents (MCIs) resulting from terrorist attacks, armed conflicts, or large-scale natural disasters can involve hundreds or thousands of victims and overwhelm local emergency services instantly~\citep{tahernejad2024application}. In these chaotic scenarios, resources are limited, communication can be disrupted, and responders are exposed to fatigue and personal risk while providing lifesaving care to multiple casualties simultaneously. The shortage of medical personnel further complicates response efforts, forcing small, exhausted teams to triage a large number of patients under hazardous conditions, where accurate patient prioritization is essential, as delays in care can be fatal. However, conventional manual triage methods can fail when faced with high casualty volumes and complex injury profiles~\cite{gabbe2022review, super1994start}. 

To address these challenges, recent efforts have focused on autonomous perception systems that improve situational awareness and decision support technologies. For example, robotic and drone platforms can be developed, capable of surveying disaster zones to identify and locate victims while remotely assessing their injuries and vital signs~\citep{agarwal2014characteristics}. These autonomous systems are equipped with multimodal sensors, such as radar, lidar, multispectral, thermal, and visible light cameras, that can noninvasively assess heart rate, respiratory rate, bleeding, amputations, responsiveness, and wounds. By providing live updates of these data to healthcare workers, these technologies can accelerate diagnosis and interventions.

Although these autonomous perception capabilities represent significant technological advances, the algorithms that power them face fundamental limitations that compromise their reliability in high-stakes MCI scenarios. Current models suffer from three critical limitations: algorithmic isolation-treating each vital sign independently, context blindness-ignoring clinical interdependencies, and brittleness under the noisy, partial data conditions prevalent to disaster scenarios. For instance, an elevated heart rate algorithm may indicate urgency, but without considering visible trauma or level of consciousness, it cannot distinguish between pain response, blood loss, or anxiety, and its distinction can determine survival. Robust decision-making requires fusing multimodal outputs and 
embedding medical expertise into the reasoning process~\citep{wadden2022defining}. 

To overcome these limitations, we propose a Bayesian Network (BN) model that integrates multiple perception algorithm outputs into a unified probabilistic estimate of the patient's condition. Rather than training on large datasets, which are hard to get, our BN is constructed from structured clinical knowledge elicited from experts. The model is designed to support inference even with missing or uncertain information, improving the consistency and interpretability of triage decisions in real time. It can probabilistically assess the likelihood of severe conditions, such as hemorrhage, based on related observations such as amputation and visible bleeding, and allows robotic systems to update assessments as new information emerges. This expert-knowledge foundation provides a robust and transparent alternative for decision support in data-scarce MCI scenarios.

In this work, we introduce a knowledge-based BN for automated triage support in MCIs.
Our model integrates vital sign estimates from perception algorithms and applies probabilistic reasoning to infer a patient's diagnosis.
It is designed for deployment in real-world scenarios where transparent decision support can directly impact survival.
The development and evaluation of this system were carried out in the context of the DARPA Triage Challenge (DTC), an initiative aimed at advancing autonomous medical triage capabilities for mass casualty incidents.

Our key contributions are (1) the first expert-elicited Bayesian Network specifically designed for MCI triage that integrates multimodal perception inputs, (2) a probabilistic reasoning framework that maintains decision transparency while handling uncertain and incomplete data, and (3) validation through realistic field scenarios conducted during the DTC Systems Competition, demonstrating that Team Chiron's integrated approach achieves significant triage accuracy gains over independent algorithmic outputs.
\section{Related work}
\label{sec:related_work}

\subsection{Bayesian networks}

Bayesian Networks (BNs)~\cite{pearl1988probabilistic}, introduced in the 1980s, are Probabilistic Graphical Models (PGMs) that represent uncertain knowledge using Directed Acyclic Graphs (DAGs), where nodes denote random variables and edges signify conditional dependencies. The power of BNs stems from their ability to compactly represent a joint probability distribution over a set of variables $\mathcal{X} = \{X_1, \dots, X_n\}$. The structure of the DAG encodes conditional independence assumptions, allowing for a computationally tractable factorization of the joint distribution:
\begin{equation}
\label{eq:pred}
    P(X_1, \dots, X_n) = \prod_{i=1}^{n} P(X_i | \text{Pa}(X_i))
\end{equation}
where $\text{Pa}(X_i)$ denotes the set of parents of variable $X_i$ in the DAG. This factorization significantly reduces the number of parameters required to define the full joint distribution, making complex probabilistic models feasible.

\subsection{Bayesian networks in medical applications}

Bayesian Networks (BNs) have been effectively used for clinical triage by modeling relationships between patient symptoms and outcomes. For instance, BN-based systems have been developed for triaging non-traumatic abdominal pain~\citep{sadeghi2006bayesian} and detecting pediatric asthma with high accuracy~\cite{sanders2006prospective}. While other studies also explore BNs for triage support, they often highlight the complexity of this approach~\citep{abad2008evolution, butcher2020extending, olszewski2003bayesian}.

Concurrently, recent advances in machine learning have enabled remote, non-contact prediction of vital signs, such as heart rate from eye movements~\citep{zheng2022heart}, SpO2 levels~\citep{al2021non}, and blood pressure via remote photoplethysmography~\citep{schrumpf2021assessment}. A recent systematic review and meta-analysis confirms the potential of this approach, showing that data-driven AI models generally outperform conventional triage tools in predicting trauma outcomes~\citep{adebayo2023exploring}. However, a significant challenge remains to translate the success of these individual models into a robust integrated system capable of assessing each casualty case as a whole. Relying solely on the isolated predictions of these diverse, often "black-box" neural networks can lead to inaccuracies, particularly in complex environments. 

\section{Methods}
\label{sec:methodology}


\subsection{Environment}
The evaluation involved three distinct MCI scenarios: \textit{Open Battlefield}, \textit{Convoy Ambush}, and \textit{Airplane Crash}.
This study focuses on the first two scenarios where the system achieved stable deployment.
Technical constraints encountered during the third scenario, which limited data acquisition for a complete probabilistic analysis, are discussed in Section~\ref{sec:results}.

Each scenario involved a timed deployment lasting up to thirty minutes, during which the robotic systems were launched into the field, navigated challenging terrain obstacles, and surveyed the environment to locate, assess and report the condition of each casualty. The emphasis was on performing rapid triage using real-time perception and inference from stand-off sensing, without human intervention or physical contact. Our system integrated data from multiple onboard perception algorithms and applied a probabilistic reasoning framework to infer physiological state and injury severity, even under degraded or incomplete sensing conditions. 

System performance was evaluated using a structured scoring framework that awarded point scores for correctly identifying physiological signs and conditions under time constraints. Each casualty presented a set of target findings, and points were awarded when the system accurately reported those findings. In Table~\ref{scoring} we present a list of physiological measurements together with their corresponding score. Additional weight was given to completing the assessment rapidly, within a "golden time window" (GW), a period where it is likely that interventions might prevent death, representing the urgency of real-world triage. The primary objective was to maximize the cumulative score for all casualties during the execution time of the scenario. 

\begin{table}[htbp]
\centering
\caption{Scoring criteria for emergency scenarios. Points are awarded based on the official DARPA Triage Challenge scoring rubric \cite{dtc_resources}. Higher scores were given for correctly assessing critical conditions, such as respiratory distress and hemorrhage, within the golden window (GW), defined as the first half of the simulation scenario, due to their urgency and clinical importance. Maximum achievable score for each casualty is equal to 12 points.}
\label{scoring}
\begin{tabular}{l l p{4cm}}
\toprule
\textbf{Field} & \textbf{Values} & \textbf{Scoring Criteria} \\
\midrule\midrule
Severe Hemorrhage    & Present / Absent & \textbf{4} if in GW \\ & & \textbf{2} if match GT \\ & & \textbf{0} otherwise \\
\midrule\midrule
Respiratory Distress & Present / Absent & \textbf{4} if in GW \\ & & \textbf{2} if match GT \\ & & \textbf{0} otherwise \\
\midrule
\midrule
Head Trauma & Wound / Normal & \textbf{2} if all match GT \\
\cmidrule{1-1}
Torso Trauma & & \textbf{1} if at least 2 match GT \\
\cmidrule{1-2}
Lower Ext. Trauma & Wound / Amputation / Normal & \textbf{0} otherwise \\
\cmidrule{1-1}
Upper Ext. Trauma & & \\
\midrule\midrule
Ocular Alertness     & Open / Closed / NT & \textbf{2} if all match GT \\
\cmidrule{1-2}
Verbal Alertness     & Normal / Abnormal / Absent / NT & \textbf{1} if at least 2 match GT \\
\cmidrule{1-1}
Motor Alertness & & \textbf{0} otherwise \\
\bottomrule

\end{tabular}
\end{table}

\subsection{System architecture and integration}

To address the multifaceted issues involved in the assessment of casualties, our solution leverages a probabilistic modeling framework grounded in domain expertise. The underlying BN model was constructed and processed using GeNIe Modeler~\citep{genie2022}. Its seamless integration into the Robot Operating System 2 (ROS2) stack was achieved through a custom interface developed with SMILE~\citep{smile2022}, which facilitates real-time inference and data exchange. This implementation directly leverages the principles outlined in~\citep{druzdzel1999smile} for practical application within a modern robotic middleware ecosystem.

The perception system was deployed on a real robotic platform equipped with a sensor suite comprising a high-resolution RGB camera, a microphone, a radar, a lidar and a thermal camera enabling the collection of visual and audio data from a stand-off distance. The system is built on the ROS2 framework, where individual physiological estimators operate as independent nodes. Each node, corresponding to an input variable in the BN model shown in Figure~\ref{fig:bn}, is an independent AI model specialized in assessing a specific physiological sign from the robot sensor feed. 

These component AI models publish their predictions to a centralized server. Crucially, the BN acts as a state estimator. When a vision node outputs a categorical label, the BN treats this as hard evidence, instantly updating the probabilities of unobserved variables. This allows the system to deduce physiological states that are not directly visible to sensors.

A core architectural principle of our system is modularity and agnosticism towards the specific perception technologies used. In the evaluated deployment, individual estimators were provided as independent "black-box" modules by separate teams. These modules published predictions without exposing internal architectures or confidence metrics, requiring our system to remain agnostic to the underlying perception technologies.

This operational constraint underscores the value of our BN-based approach. The framework is designed to robustly fuse outputs from heterogeneous sources with unknown and potentially variable performance characteristics. By probabilistically reasoning over these inputs, the BN can mitigate the impact of noise or failure in any single module, improving the overall reliability of the final assessment.

Upon receiving inputs, the BN performs the inference process detailed in Equation \ref{eq:pred} to estimate unobserved physiological variables and generate a probabilistic assessment of the patient's condition. This structure enables reasoning over incomplete or noisy inputs and supports on-the-fly updates as new data become available. The core strength of the system lies in its modularity. Because each physiological estimator operates independently, the system can incorporate new algorithms or sensing modalities with minimal architectural changes. This modularity ensures that the system remains adaptable to evolving technologies in emergency response.

The parameterization of the Bayesian Network, specifically the definition of the Conditional Probability Tables (CPTs) for each node, was carried out through a knowledge-driven engineering process. Due to the absence of publicly available standardized datasets for MCI scenarios, a data-driven approach was not feasible. Instead, the CPTs were populated based on qualitative rules and heuristics derived from consultations with medical experts, as detailed in Section~\ref{subsec:3.3_expert}. This process involved translating qualitative statements (e.g., ``highly likely,'' ``possible,'' ``unlikely'') into quantitative probability values.

To ensure consistency and facilitate replicability, we adopted the following convention:
\begin{itemize}
    \item \textbf{Strong causal relationships} (e.g., a lower extremity amputation almost always causes severe hemorrhage) were modeled using probabilities in the range of 0.8 -- 0.95.
    \item \textbf{Moderate correlations} (e.g., head trauma may be associated with abnormal verbal alertness) were assigned values in the 0.4 -- 0.6 range.
    \item \textbf{Weak or rare dependencies} were represented by low probabilities, close to the prior probabilities of the baseline.
\end{itemize}

For instance, the probability for "Closed" for "Ocular Alertness" node in the case when prior of "Head Trauma" was "Wound" was set to 0.7. Although exact values were heuristically assigned, this structured methodology provides transparency and a framework for future model calibration should relevant data become available.

The system is lightweight (under 100MB) and operates with inference times below 1ms even on a legacy hardware such as Raspberry Pi 3.

\subsection{Expert knowledge elicitation process}
\label{subsec:3.3_expert}
\begin{figure}[th]
    \centering
    \caption{Bayesian network architecture illustrating the relationships among vital signs. Arrows represent causal dependencies derived from expert knowledge. During inference, the flow of information is bidirectional: observing a symptom (child node) updates the probability of the underlying cause (parent node).}
    \hspace{0.02cm}
    \includegraphics[width=1\linewidth]{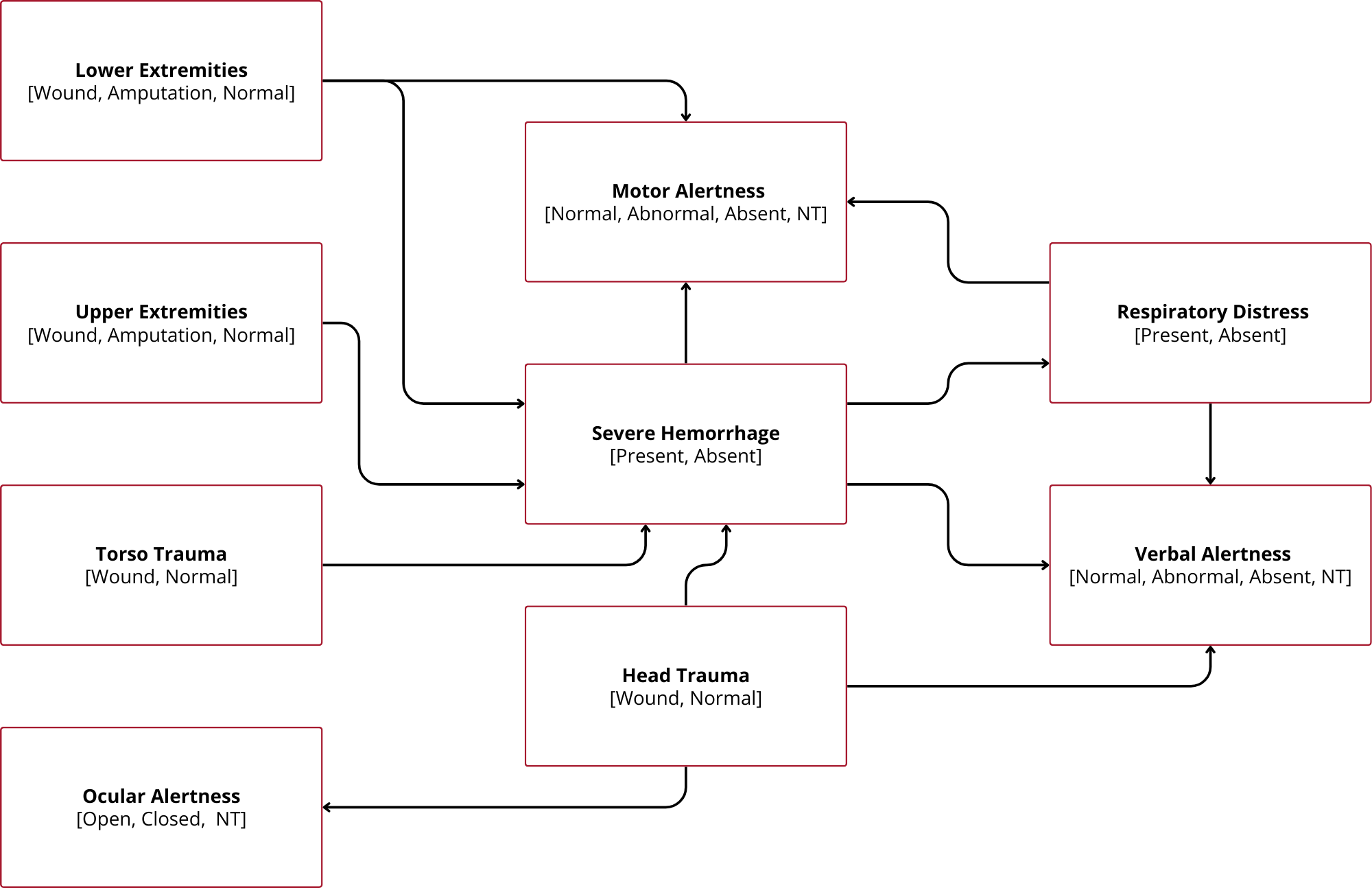}

   \label{fig:bn}
\end{figure}

The design and parameterization of the BN, whose structure is shown in Figure~\ref{fig:bn}, involved three domain experts, each with more than a decade of field experience in emergency medicine and rapid response medical interventions.

The methodology consisted of an iterative cycle. Initially, our engineering team conducted a series of individual interviews to identify the most critical observable indicators and their high-level relationships. Based on this qualitative input, an initial BN structure and preliminary CPTs were drafted. These drafts translated qualitative expert statements (e.g., "a lower extremity amputation almost always causes severe hemorrhage") into quantitative probability values.

In the second phase, the drafted model was presented back to the experts in follow-up sessions. Their role was to validate the model logic, identify potential gaps, and challenge assumptions. This feedback loop allowed the engineering team to manually refine the CPTs to ensure that the model's inferential behavior was balanced and consistent with established clinical reasoning. This expert-in-the-loop approach, although not fully automated, ensured that the final model was computationally robust and grounded in real-world medical expertise.

\section{Results}
\label{sec:results}

The effectiveness of our framework was evaluated during the DTC Round 1 Systems Competition.
We report performance results for the \textit{Open Battlefield} and \textit{Convoy Ambush} scenarios. The third mission, \textit{Airplane Crash}, is excluded from this analysis due to a hardware-level failure prior to deployment that prevented consistent data acquisition. Despite this, Team Chiron’s performance in the remaining missions was sufficient to secure 4th place out of 11 teams overall.

Despite these technical challenges in the final run, the performance of Team Chiron in the first two scenarios was sufficient to secure 4th place out of 11 teams overall. This result demonstrates the robustness of our Bayesian reasoning framework when integrated with an operational perception pipeline.

The evaluation involved 20 distinct casualty cases, portrayed by both human volunteers and advanced medical mannequins. These scenarios designed by DARPA tested the system's performance under significant environmental and tactical challenges, confronting the robotic system with real-world sensor noise and physical complexity.

In both scenarios, we compared the triage assessment scores achieved by the base perception system operating alone (``Robot Score'') against the integrated system featuring our Bayesian Network (``Robot + BN Score''). The baseline ``Robot Score`` system consists of independent nodes reporting without a central reasoning layer or validation mechanism. Points were awarded by DTC technical observers based on correctly identifying key vital signs for each casualty. The scoring framework emphasized both accuracy and timeliness, with greater weight given to the early detection of life-threatening conditions. The comparative results for both scenarios are detailed in Table~\ref{tab:performance_combined}.

Our findings demonstrate a substantial and consistent performance improvement when the BN is integrated with the autonomous triage system. Across both scenarios, the BN significantly improved the system's ability to reason from the incomplete or uncertain sensor data that characterized these field tests. As shown in Table~\ref{tab:performance_combined}, the integration of the BN more than doubled the robot's score in \textit{Open Battlefield} scenario, from 25 to 61 of the total possible points, and nearly tripled it in \textit{Convoy Ambush} scenario, from 16 to 45. 

Table~\ref{tab:performance3} further details this improvement by breaking down the results into \textbf{Correct Assignments} (assessments matching ground truth) and \textbf{Assignment Attempts} (any instance where the system provided an output). System \textbf{Reliability}, defined as the ratio of completed measurements to the total possible assessments (180), increased from 0.31 to \textbf{0.95}. Furthermore, \textbf{Accuracy} (correct assignments among attempts) improved from 46\% to 56\%. Notably, the integration of the BN led to a substantial increase in overall \textbf{Performance}, raising the correct assignment rate across all potential vitals from 14\% to \textbf{53\%}. The low baseline performance was often caused by transient sensor failures, a gap effectively bridged by the BN's inference capabilities. This 3.84-fold increase in correctly assigned vitals highlights the BN's critical role in transforming fragmented perceptual outputs into a coherent assessment, particularly when individual vision-based algorithms fail due to environmental occlusions or other real-world challenges.

\begin{table}[htbp]
\centering
\caption{Comparison of the system performance with and without BN}
\label{tab:performance_combined}
\begin{tabular}{c c c c c}
\toprule
& \multicolumn{2}{c}{\textbf{Open Battlefield}} & \multicolumn{2}{c}{\textbf{Convoy Ambush}} \\
\cmidrule(lr){2-3} \cmidrule(lr){4-5}
\textbf{Casualty Id} & \textbf{Robot Score} & \textbf{Robot + BN Score} & \textbf{Robot Score} & \textbf{Robot + BN Score} \\
\midrule\midrule
1  & 9 & 9 & 0 & 0 \\
2  & 0 & 3 & 0 & 3 \\
3  & 0 & 3 & 2 & 7 \\
4  & 0 & 3 & 2 & 7 \\
5  & 0 & 3 & 0 & 4 \\
6  & 0 & 5 & 1 & 3 \\
7  & 4 & 5 & 8 & 11 \\
8  & 0 & 7 & 1 & 3 \\
9  & 8 & 9 & 2 & 7 \\
10 & 4 & 7 & N/A & N/A \\
11 & 0 & 7 & N/A & N/A \\
\midrule\midrule
\textbf{Total}       & 25/132 & \textbf{61/132} & 16/108 & \textbf{45/108} \\
\bottomrule
\end{tabular}
\end{table}

\begin{table}[htbp]
\centering
\caption{Comparison of Correct Assignments and Assignment Attempts with and without BN across all casualties.}
\label{tab:performance3}
\begin{tabular}{l c c c c}
\toprule
& \multicolumn{2}{c}{\textbf{Correct Assignments}} & \multicolumn{2}{c}{\textbf{Assignment Attempts}} \\
\cmidrule(lr){2-3} \cmidrule(lr){4-5}
\textbf{Vital} & \textbf{Robot} & \textbf{Robot + BN} & \textbf{Robot} & \textbf{Robot + BN} \\
\midrule\midrule
Severe Hemorrhage     & 6 & 12 & 12 & 19 \\
Respiratory Distress  & 5 & 16 & 6 & 19 \\
Head Trauma           & 0 & 15  & 0 & 19 \\
Torso Trauma          & 0 & 11  & 0 & 19 \\
Lower Ext. Trauma     & 8 & 11 & 14 & 19 \\
Upper Ext. Trauma     & 3 & 8  & 14 & 19 \\
Motor Alertness       & 0 & 8  & 0  & 19 \\
Verbal Alertness      & 3 & 7  & 8  & 19 \\
Ocular Alertness      & 0 & 8  & 1  & 19 \\
\midrule\midrule
\textbf{Total}        & 25 & 96 & 55 & 171 \\
\cmidrule(lr){2-3}\cmidrule(lr){4-5}
\textbf{Reliability} & N/A & N/A & 0.31 & \textbf{0.95} \\
\cmidrule(lr){2-3}\cmidrule(lr){4-5}
\textbf{Performance} & 14\% & \textbf{53\%} & N/A & N/A \\
\cmidrule(lr){2-3}\cmidrule(lr){4-5}
\textbf{Accuracy} & 46\% & \textbf{56\%} & N/A & N/A \\
\bottomrule
\end{tabular}
\end{table}

\section{Discussion}
\label{sec:discussion}

The presented approach to the fusion of casualty assessment components demonstrates a substantial improvement in reliability and accuracy over the baseline system. The observed gains highlight the profound impact of explicitly modeling uncertainty and inter-modal dependencies, a capability currently lacking in standard vision-based triage tools. During the DTC field evaluations, the system's performance improved as additional evidence became available, emulating the iterative reasoning of human medics at machine speeds. This shift from static, one-off assessments to continuous probabilistic reasoning is a key step forward toward effective autonomous triage systems in high-stakes environments.

Our experience in the challenge underscores the insufficiency of using isolated "black box" models. As Team Chiron utilized various independent perception modules, we observed that lacking domain-level integration can lead to inconsistent or implausible predictions. Our choice of a Bayesian Network offered a critical advantage by enabling modularity and extensibility, allowing us to fuse heterogeneous outputs from different sub-teams into a coherent assessment without needing access to the internal architectures of the individual AI estimators.

The proposed architecture ensured graceful degradation when data were incomplete—a frequent occurrence in the DTC scenarios due to sensor occlusions, dust, or environmental noise. The failure of a single input module did not cause a catastrophic failure of the entire triage assessment; instead, the BN marginalized over the missing variables, providing the best possible diagnostic estimate from the remaining functional sensors. This fault tolerance proved to be a fundamental requirement for operational success in the field.

The probabilistic framework is inherently extensible. Although the current implementation treats categorical inputs as direct evidence, the BN structure is designed to accommodate future enhancements, such as incorporating model confidence scores or rejecting low-probability inputs. This forward-looking flexibility is essential for future phases of the DTC, where more complex injury profiles and higher levels of environmental degradation are expected.

Even though our initial results from Round 1 are encouraging, a more comprehensive evaluation is necessary. Future work will involve direct comparisons with other hybrid architectures and further refinement of the expert-defined CPTs based on the data collected during these field trials.
\section{Conclusion and future work}
\label{sec:conclusion}

We presented a Bayesian Network-based decision support framework that fuses outputs from multiple computer vision models to automate trauma assessment in mass casualty incidents. By leveraging expert-defined clinical rules, our model provides a robust, transparent, and training-data-independent solution for high-stakes triage. The empirical validation conducted during the DTC Round 1 demonstrated that our integrated system substantially improves the accuracy and diagnostic coverage of casualty assessments compared to vision-only baselines.

Based on the challenges identified during our participation in the DTC, we propose the following directions for future development.

\subsection{Robustness and sensor degradation}
Our field experiments underscored the critical importance of robustness testing in degraded environments (smoke, dust, and debris). While the BN mitigates cascading error propagation, future iterations of Team Chiron's platform will focus on tighter integration of sensor confidence metrics. This will allow the BN to dynamically adjust the weighting of inputs from specific modalities (e.g., thermal vs. RGB) when environmental conditions severely degrade.

\subsection{Human-AI collaboration and tactical optimization}
The current system frees up a responder's cognitive resources by synthesizing complex data into a clear diagnostic picture. We envision extending this approach to optimize the scheduling of care for multiple casualties. In the context of DTC's evolving requirements, we plan to incorporate health trajectory forecasting, allowing the system to not only assess the current state but also predict the urgency of intervention. This will require specialized training for emergency responders to effectively trust and collaborate with the autonomous system in the field.

\subsection{Data Quality and Field Data Acquisition}
A fundamental challenge remains the lack of standardized MCI datasets for machine learning. The physical rounds of the DTC provide a unique and invaluable opportunity to collect high-fidelity, annotated data from realistic scenarios. We are currently processing the data captured during Round 1 to refine our CPTs and we encourage the research community to join us in establishing open-access datasets for autonomous triage.

\subsection{Ethical Considerations and Safeguards}
Relying on AI-driven decision support in life-or-death scenarios raises profound ethical questions. Our future development will prioritize "human-in-the-loop" mechanisms and robust safeguards to ensure that the system remains a supportive tool for expert responders, rather than a replacement for human judgment in critical care.

\begin{ack}
This work has been partially supported by the Defense Advance Research Projects Agency (award HR00112420329) and National Science Foundation (awards 2427948 and 2406231). The views, opinions, and/or findings expressed are those of the authors and should not be interpreted as representing the official views or policies of the Department of Defense or the U.S. Government. We also thank the DARPA Triage Challenge organizers and the technical observers for their logistical support during the field evaluations.
\end{ack}

\bibliographystyle{plainnat}
\bibliography{main}

\end{document}